# Refined Infrared Small Target Detection Scheme with Single-Point Supervision

Jinmiao Zhao[1,2,3,4], Zelin Shi[1, 2, *], Chuang Yu[1,2,3,4], and Yunpeng Liu[1, 2]

[1]Key Laboratory of Opto-Electronic Information Processing, Chinese Academy of Sciences
[2]Shenyang Institute of Automation, Chinese Academy of Sciences
[3]Institutes for Robotics and Intelligent Manufacturing, Chinese Academy of Sciences
[4]University of Chinese Academy of Sciences
{zhaojinmiao, zlshi, yuchuang, ypliu}@sia.cn

**Abstract.** Recently, infrared small target detection with single-point supervision has attracted extensive attention. However, the detection accuracy of existing methods has difficulty meeting actual needs. Therefore, we propose an innovative refined infrared small target detection scheme with single-point supervision, which has excellent segmentation accuracy and detection rate. Specifically, we introduce label evolution with single point supervision (LESPS) framework and explore the performance of various excellent infrared small target detection networks based on this framework. Meanwhile, to improve the comprehensive performance, we construct a complete post-processing strategy. On the one hand, to improve the segmentation accuracy, we use a combination of test-time augmentation (TTA) and conditional random field (CRF) for post-processing. On the other hand, to improve the detection rate, we introduce an adjustable sensitivity (AS) strategy for post-processing, which fully considers the advantages of multiple detection results and reasonably adds some areas with low confidence to the fine segmentation image in the form of centroid points. In addition, to further improve the performance and explore the characteristics of this task, on the one hand, we construct and find that a multi-stage loss is helpful for fine-grained detection. On the other hand, we find that a reasonable sliding window cropping strategy for test samples has better performance for actual multi-size samples. Extensive experimental results show that the proposed scheme achieves state-of-the-art (SOTA) performance. Notably, the proposed scheme won the third place in the "ICPR 2024 Resource-Limited Infrared Small Target Detection Challenge Track 1: Weakly Supervised Infrared Small Target Detection".

## 1    Introduction

With the continuous development of infrared imaging technology and digital image processing technology, infrared small target detection technology has been widely used in many fields, including maritime surveillance, and early warning systems. [1-3]. Existing research generally focuses on infrared small target detection with full supervision, whereas research on infrared small targets with single-point supervision is very rare. However, due to the small size of infrared small targets and the lack of target intrinsic features, it is difficult and time-consuming to directly manually annotate fully supervised samples. Therefore, infrared small target detection with single-point supervision is extremely challenging and meaningful.

Early research on infrared small target detection is based on non-deep learning methods, such as local contrast measure (LCM) [4], multi-scale local contrast measure (MLCM) [5], and infrared image patch model (IPI) [6]. However, this type of model-driven method has excellent detection performance for specific scenarios, but often results in many false detections when the scenarios are complex and diverse. Different from non-deep learning methods, deep learning methods are data-driven methods that aim to automatically extract features through input data and construct a mapping from input data to output data. This type of method has better detection effects in complex scenes, such as ALCL-Net [7], DNA-Net [8], and MSDA-Net [9]. However, the above networks are all designed for infrared small target detection tasks with full supervision, and it will be difficult to effectively migrate directly to single-point supervision tasks. Therefore, Ying et al. [10] propose a label evolution with single point supervision (LESPS) framework, which can apply the fully supervised infrared small target detection network to single-point supervision tasks. However, we find that direct migration suffers from low segmentation accuracy and detection rates.

Therefore, we aim to build a refined infrared small target detection scheme with single-point supervision. First, network performance is crucial to the detection effect of the entire scheme. Therefore, we explore several excellent infrared small target detection methods based on the LESPS framework and select the high-performance GGL-Net [11] as our backbone network. Secondly, to improve the comprehensive performance of the scheme, we construct a complete set of post-processing strategies. On the one hand, to improve the segmentation accuracy, we use a combination of test-time augmentation (TTA) and conditional random field (CRF) [12] for post-processing. This post-processing can reduce false detections and missed detections, enhance the connectivity of detection results, and improve the accuracy of target edges. On the other hand, we introduce the adjustable sensitivity (AS) strategy [13] for post-processing. It takes full account of the advantages of multiple detection results and reasonably adds some areas with low confidence into the fine segmentation image in the form of centroid points to significantly improve the detection rate. In addition, to further improve the detection performance and explore the characteristics of this task, we explore the loss function and the test image input strategy. On the one hand, we consider that the single-stage loss cannot play a good role in constraining the network for this task. Therefore, we construct a multi-stage loss to strengthen the network constraints, promote the gradual refinement of predictions and enhance the robustness of the model. On the other hand, we consider that directly inputting the original image or resizing the original image to a fixed size input will have limitations. The former will have equipment limitations, and the latter will result in a smaller target area that is more difficult to detect. Therefore, we construct a sliding window cropping strategy, which achieves fine detection by cropping the test images into fixed sizes in turn, detecting them in batches, and then restoring them in reverse.

In summary, we construct an innovative refined infrared small target detection scheme with single-point supervision, which has excellent segmentation accuracy and a high detection rate. Our code can be available online[1]. Notably, the proposed scheme won the third place in the "ICPR 2024 Resource-Limited Infrared Small Target Detection Challenge Track 1: Weakly Supervised Infrared Small Target Detection". The contributions of this manuscript can be summarized as follows:

(1) We introduce the LESPS framework and explore the performance of various excellent infrared small target detection networks based on this framework.

(2) To improve the comprehensive performance, we propose a complete set of post-processing strategies. On the one hand, to improve the segmentation accuracy, we use the TTA and CRF for post-processing. On the other hand, to improve the detection rate, we introduce an AS strategy, which reasonably adds some targets with low confidence into the fine segmentation image in the form of centroid points.

(3) To further improve the detection performance and explore the characteristics of this task, on the one hand, we explore the loss function. We construct and find that multi-stage loss constraints are helpful for refined detection. On the other hand, we explore the test image input strategy and find that properly cropping the test samples during testing has better performance.

## 2   Related Work

### 2.1   Non-deep learning-based infrared small target detection methods

Early infrared small target detection mainly used model-driven non-deep learning methods, which can be divided into three categories: background suppression methods, human visual system simulation methods, and image data structure analysis methods. Background suppression-based methods mainly highlight the target by reducing background noise. These methods include the Max-Median filter [14], Top-Hat filter [15], bilateral filter [16], and two-dimensional minimum mean square error filter [17]. Human visual system-based methods enhance salient areas in images by simulating human visual characteristics, thereby improving the target detection rate. These methods include the local contrast measure (LCM) [4], multi-scale local contrast measure (MLCM) [5] and weighted local difference measure (WLDM) [18]. The image data structure-based method transforms the small target detection problem into an optimization problem of low-rank and sparse matrix recovery. This type of method detects targets by analyzing and utilizing the structural characteristics of images. These include the infrared image patch model (IPI) [6], the non-negative infrared patch-image model based on partial sum minimization of singular values (NIPPS) [19], and the minimum joint L2,1 norm (NRAM) [20]. The above methods have excellent performance in specific scenarios. However, in highly complex real-world scenarios, such as when images contain high-brightness noise, these model-driven methods often perform poorly and are prone to false detection.

---

[1] https://github.com/YuChuang1205/Refined-IRSTD-Scheme-with-Single-Point-Supervision

## 2.2 Deep learning-based infrared small target detection methods

Different from the traditional non-deep learning-based infrared small target detection method, the deep learning-based methods adopt a data-driven approach to identify and learn the characteristics of infrared small targets. In the early days, Liu et al. [21] design a multi-layer perception network for infrared small target detection and show excellent detection performance. Subsequently, Dai et al. propose the asymmetric context module [22] and the attention local contrast network (ALCNet) [23] to integrate the model-driven method with the data-driven method. To further enhance the extraction of the relationships among local features, Yu et al. successively construct a multi-scale local contrast learning network (MLCL-Net) [24] and an attention-based local contrast learning network (ALCL-Net) [7]. Recently, Li et al. propose an innovative densely nested attention network (DNANet) [8] based on UNet++. DNANet achieves repeated fusion and enhancement of features by introducing densely nested interaction modules and channel space attention modules. Wu et al. also proposed an improved U-Net nested network (UIU-Net) [25], which focuses on the fusion of multi-scale features. UIU-Net uses multi-scale feature fusion technology to enhance the model's ability to process targets of different scales. Subsequently, Zhao et al. consider the importance of gradient information and directional features in infrared small target detection and propose GGL-Net [11] and MSDA-Net [9] respectively. GGL-Net focuses on capturing image gradient information and can more accurately identify the edges and detail features of targets by introducing a gradient-guided learning mechanism. MSDA-Net emphasizes the extraction of directional features and enhances the detection capability of small targets in complex backgrounds by fusing multi-scale directional features. In addition, Zhang et al. propose an infrared shape network (ISNet) [26], which focuses on extracting the shape features of targets. Although the above infrared small target detection method has achieved excellent performance under fully supervised tasks, directly applying it to single-point supervision tasks will result in significant performance degradation. Therefore, Ying et al. [10] propose a label evolution framework called label evolution with single point supervision (LESPS). This framework can be used to migrate the infrared small target detection method with full supervision to the single-point supervision task. However, the effect after direct migration is still insufficient and needs to be improved. Therefore, we intend to build a refined infrared small target detection scheme with single-point supervision.

## 3 Method

### 3.1 Overall scheme introduction

To achieve accurate detection of infrared small targets under single-point supervision, we propose a refined detection scheme. From Fig. 1, the proposed refined detection scheme can be divided into two parts: a training phase and an inference phase. In the training phase, first, the original training image is randomly cropped into fixed-size image patches (256×256 pixels). This process can generate more training samples. Then, the cropped image patches are input into the network for training after data augmentation. Since the true labels of this task are point labels, we use the label evolution with single point supervision (LESPS) framework, which generates refined pseudo labels in each epoch to achieve accurate target detection gradually. Based on the LESPS framework, we explore a variety of excellent infrared small target detection networks and finally use the GGL-Net proposed in our previous work [11] as the backbone network. In addition, to further improve the detection accuracy, we explore the loss function and construct a multi-stage focal loss for GGL-Net to promote network forward optimization. In the inference phase, first, the original test image is generated into horizontally flipped and vertically flipped images through the test-time augmentation (TTA) strategy. Second, the image is cropped through a sliding window, and the image is cropped into multiple image blocks in turn. The area that is less than a window is filled with black areas. Third, the above image patches are input into the trained model for inference, and the detection results are output. Fourth, the detection results of the image patches are sequentially spliced and flipped for restoration. At the same time, the three results of the same sample are image fused. In addition, for the fused segmentation result, we also perform effective area cropping to remove the black edges previously added. Finally, to further improve the overall performance of infrared small target detection, we continue to perform conditional random field (CRF) and adjustable sensitivity (AS) strategies [13] on the segmentation map output in the previous step for post-processing to improve the segmentation accuracy and detection rate, respectively.

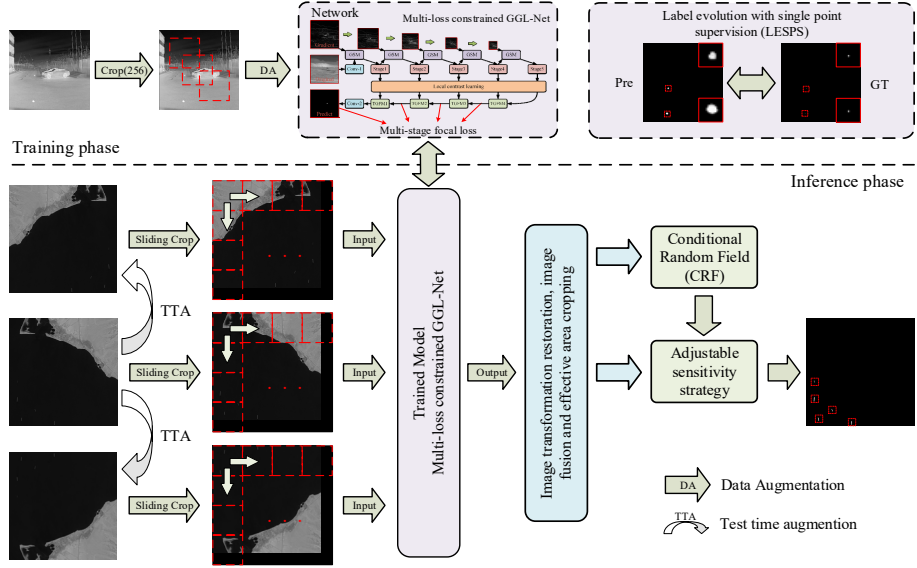

**Fig. 1.** Overall structure of the proposed scheme.

## 3.2 GGL-Net

An excellent network is crucial to the entire refined infrared small target detection scheme, and it greatly affects the final detection accuracy. Combining the characteristics of single-point supervision tasks, we explore a variety of excellent infrared small target detection networks based on the LESPS framework. Based on comprehensive comparison, we finally use the GGL-Net proposed in our previous work [11] as the backbone network. The network structure of GGL-Net is shown in Fig. 1. Different from other infrared small target detection networks that focus on the study of network architecture, GGL-Net uses the proposed gradient supplementation module to input gradient amplitude images of different scales into the main branch to guide the training optimization of the network via gradient information. At the same time, GGL-Net also proposes and uses a bidirectional guided fusion module (TGFM), which fully considers the characteristics of feature maps at different levels and effectively extracts richer semantic information and detail information through bidirectional guidance. GGL-Net achieves excellent detection performance for infrared small target detection with full supervision. Considering that this task is single-point supervision rather than full supervision, we explore its loss function and use a multi-stage loss, as detailed in Section 3.5.

## 3.3 Post-processing strategy

To further improve the comprehensive performance of infrared small target detection, we construct a complete post-processing strategy. Specifically, to improve the segmentation accuracy, we use the TTA strategy and the CRF strategy successively. For the TTA strategy, we move the target area in the image by flipping the test image and then restore and fuse the image after sequential detection through the network. When detecting images, the deep learning network combines local detail information and overall semantic information for judgment. The offset of the target position will cause the network to produce a new inference result for the image. Reasonable fusion of these inference results helps reduce false detection and missed detection and improves the accuracy of target edges. For the CRF strategy, it takes into account the lack of complete mask labels for infrared small target detection with point supervision, and the detection performance of the network is lower than that of fully supervised learning. Therefore, we introduce a CRF strategy, which helps to enhance the connectivity of the detection results and improve the accuracy of the target edges. An excellent infrared small target detection method needs to have a high detection rate in addition to high segmentation accuracy. To improve the detection rate, we used the adjustable sensitivity (AS) strategy [13]. This strategy introduces the concepts of strong targets and weak targets. For strong targets, the AS strategy fully retains them to ensure high segmentation accuracy. For weak targets, the AS strategy adds them to the finely segmented binary image in the form of centroid points to facilitate the detection of difficult targets and significantly improve the detection rate. The original AS strategy is to set two thresholds for a single output probability map to divide strong and weak target areas. In this scheme, we observe that the CRF strategy will shave off the low-confidence areas (weak target areas) in the generated mask image to avoid false detection and ensure high segmentation accuracy. Therefore, we set the feature map after the CRF strategy in the test phase as the strong target area (finely segmented binary map). Then, the threshold of the feature map before CRF strategy processing is adjusted to generate a weak target area and added to the fine binary map after CRF strategy processing in the form of centroid points to significantly improve the detection rate. That is, the processing of a single feature map in the original AS strategy is upgraded to the fusion processing of a pair of feature maps.

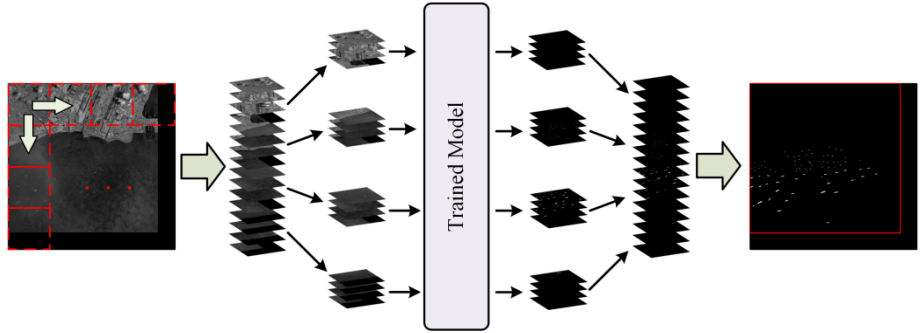

**Fig. 2.** Sliding window cropping strategy

### 3.4 Sliding window cropping strategy

The infrared small target images actually collected may be of multiple sizes. When some large-size images are directly input into the network for detection, it is easy to address the problems of insufficient device resources and inability to calculate. At the same time, if the image is directly resized to a fixed size, since infrared small targets already have small sizes, directly reducing a large-size image to a fixed size will result in a smaller target size, making detection more difficult. Therefore, we use a sliding window cropping strategy. From Fig. 2, the strategy first crops the test samples into image patches according to a fixed window size and combines them into a multidimensional array. The area that is less than the size of a window is filled with black areas. Second, the multidimensional array is input into the trained network in batches for detection, and the corresponding segmentation results are output. This process can make full use of the parallel processing of the GPU to accelerate the inference process. Then, the segmentation results in batches are sequentially combined and restored via sequential splicing operations. Finally, the supplementary black edge area is removed to extract the valid segmentation area. Since the image size input to the network during training is a fixed size cropped from the original sample, using the same size input during testing has excellent detection accuracy.

### 3.5 Multi-stage loss

To facilitate forward optimization of the network, we construct a multi-stage loss for GGL-Net. Similar to the previous single-stage focal loss, we add a convolutional layer to each of the multi-stage outputs and upsample them to the corresponding output size for loss calculation. The use of multi-stage losses can achieve gradual refinement of detection results, enhance the robustness of the model, and better represent features. The formula is as follows:

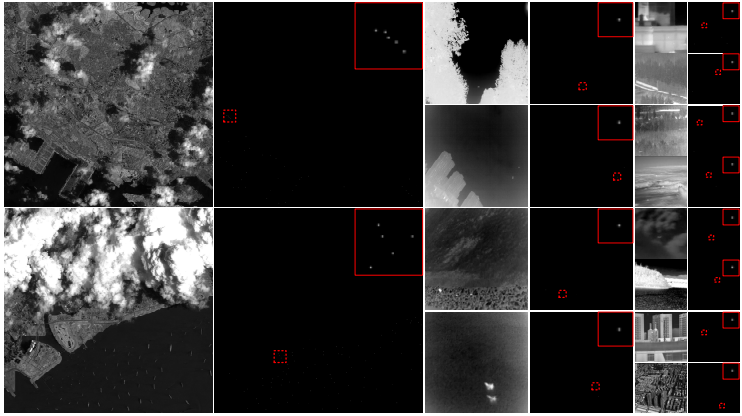

**Fig. 3.** Some samples from the WP-IRSTD dataset and their point labels.

$$L_{all} = FL_1 + FL_2 + FL_3 + FL_4 \qquad (1)$$

$$FL(p_t) = -\sum_{i=1}^{N} \alpha_t (1-p_t)^\gamma \log(p_t) \qquad (2)$$

where $L_{all}$ denotes the total loss. $FL_1$, $FL_2$, $FL_3$ and $FL_4$ denote the losses at each stage. $p_t$ denotes the predicted probability of the correct category. $\alpha_t$ is the balancing factor, which is used to balance the weights between different categories. $\gamma$ is a regulation factor used to adjust the attention paid to difficult-to-classify samples. The multi-stage loss strengthens the optimization constraints on the network, promotes the network to refine feature representation layer by layer, and extracts richer hierarchical information.

## 4 Experiments

### 4.1 Dataset

For the datasets studied, we use seven public datasets, SIRST-V2, IRSTD-1K, IRDST, NUDT-SIRST, NUDT-SIRST- sea, NUDT-MIRSDT, and Anti-UAV410, totaling 9,000 images, as experimental datasets. The images in this dataset come from multiple observation perspectives (land-based, air-based, and space-based), contain multiple target types (extended targets, spot targets, and point targets), cover multiple bands (shortwave, longwave, near-infrared), and have multiple resolutions (such as 256×256, 512×512, and 1024×1024). Considering that the true label of the dataset is a complete label, we generate the point label corresponding to each mask label as the label for the final network training. Considering its characteristics, we call the 9,000-image dataset the wide-area point label infrared small target detection (WP-IRSTD) dataset. Fig. 3 shows some samples from the WP-IRSTD dataset. We can see that there are significant differences between samples and that the target types are quite diverse. In the experiment, we divided 9000 images into atraining set and a test set at a ratio of 5:1. The samples in the test set are not converted into point labels.

**Table 1.** Performance comparison of various methods on the WP-IRSTD dataset.

| Methods | IoU | nIoU | $P_d$ | $F_a$ (×10$^{-6}$) | *Score* |
|---|---|---|---|---|---|
| ACM2[22] | 22.34 | 14.51 | 39.78 | 65.12 | 31.06 |
| ALCNet [23] | 26.79 | 17.39 | 53.31 | 64.48 | 40.05 |
| MLCL-Net [24] | 29.56 | 20.32 | 52.27 | 49.35 | 40.92 |
| ALCL-Net [7] | 38.70 | 23.34 | 53.10 | 28.48 | 45.90 |
| DNANet [8] | 28.86 | 22.38 | 53.08 | 32.48 | 40.97 |
| GGL-Net [11] | **39.50** | **28.13** | **59.31** | **25.71** | **49.41** |

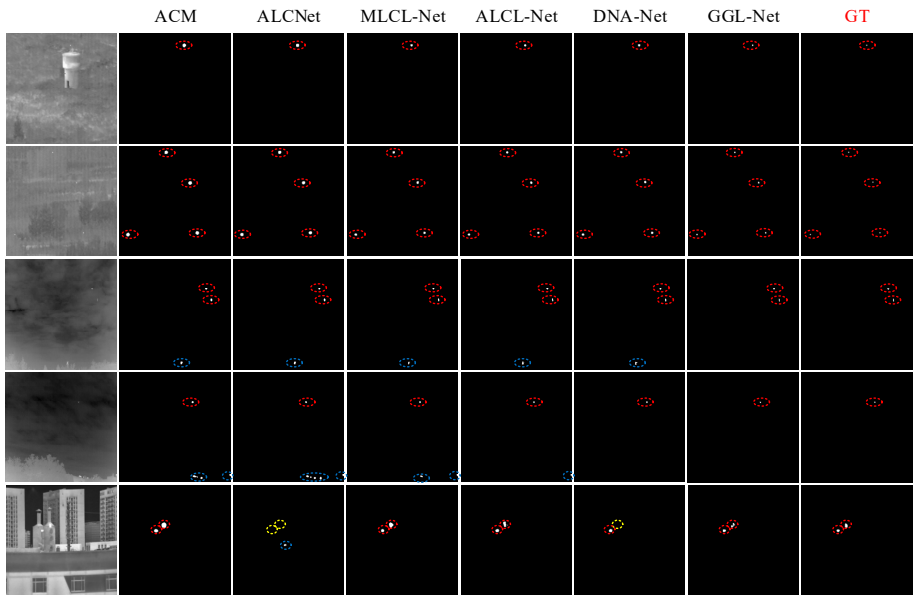

**Fig. 4.** Comparison of some detection results of various methods on the WP-IRSTD dataset. Red circles denote correct detections. Blue circles denote false detections. Yellow circles denote missed detections.

### 4.2 Experimental settings

For the experimental environment and parameter settings, the operating system is Ubuntu 18.04, and the GPU is RTX 3090 24 GB. The number of epochs epoch is 300, and the training batch size is 16. The test batch size is 1, and the number of batch image patches used during testing is 32. The learning rate is 5e$^{-4}$. A cropping window of 256 × 256 pixels is used for both training and testing.

For the evaluation metrics, consistent with "ICPR 2024 Resource-Limited Infrared Small Target Detection Challenge Track 1: Weakly Supervised Infrared Small Target Detection", we use pixel-level metrics (intersection over union (IoU), normalized IoU (nIoU)) and target-level metrics (detection rate $P_d$ and false alarm rate $F_a$) to evaluate the detection accuracy of the network. Consistent with the competition, the formula for the network's final performance evaluation metric *Score* is as follows:

**Table 2.** Performance comparison of single-stage loss and multi-stage loss

| Scheme | IoU | nIoU | $P_d$ | $F_a (\times 10^{-6})$ | **Score** |
|---|---|---|---|---|---|
| Single-stage loss | 39.50 | 28.13 | 59.31 | 25.71 | 49.41 |
| Multi-stage loss | **43.94** | **31.04** | **56.51** | **17.41** | **50.22** |

$$Score = \alpha \times IoU + (1-\alpha) \times P_d \qquad (3)$$

where $\alpha = 0.5$. The IoU reflects the segmentation accuracy of the network, and the $P_d$ reflects the network's ability to detect small targets. In addition, $F_a$ must be less than $1e^{-4}$ for the final result to be considered valid.

### 4.3 The selection of backbone networks

To improve the performance of infrared small target detection with single-point supervision, we explore several excellent infrared small target detection methods under the LESPS framework. The experimental results are shown in Table 1. From the experimental results, we can see that GGL-Net has achieved the best performance advantage. Specifically, compared with the results of MLCL-Net, ALCL-Net and DNANet, GGL-Net improves the IoU, $P_d$ and *Score* by 0.8-10.64, 6.21-7.04, and 3.51-8.49 respectively. To more intuitively demonstrate the performance of various excellent methods, we have visualize some of the results. From Fig. 4, the mask generated by GGL-Net is more refined. The small target areas detected by the other methods are larger than the true labels. At the same time, compared with other methods, GGL-Net results in fewer false detections and missed detections. Therefore, we choose GGL-Net as our backbone network.

### 4.4 Effect verification of the multi-stage loss

To further improve the network performance, we compare the performance of single-stage loss and multi-stage loss based on the backbone network GGL-Net. From Table 2, Compared with single-stage loss, using multi-stage loss can improve the *Score* by 1.64% (from 49.41 to 50.22). The use of multi-stage loss can effectively guide the network to gradually refine the characteristics of the target and improve the precision of infrared small target segmentation.

### 4.5 Effect verification of the adjustable sensitivity strategy

To verify the impact of the AS strategy on our scheme, we use GGL-Net as the backbone model and add multi-stage loss for constraints. The sliding window size during testing is 256×256. The TTA and CRF strategies are used for post-processing in the experiment. The specific experimental results are shown in Tables 3 and 4.

There are two thresholds in the AS strategy used in our proposed scheme that need to be experimentally explored, namely, the threshold (th1) at which the image becomes a binary image after CRF processing and the threshold (th2) at which weak target information is to be extracted from the image before CRF processing. To further verify the effect of threshold th1 in the AS strategy, we adjust the results of th1 without using threshold th2. From Table 3, when th1 is equal to 0.005, compared with the traditional setting of 0.5, the final *Score* is improved by 9.77 (from 45.14 to 54.91). When th1 is equal to 0.01, compared with the initial th1 equal to 0.5, the IoU,

**Table 3.** Effect verification of the threshold th1 in the AS strategy.

| th1 | IoU | nIoU | $P_d$ | $F_a (\times 10^{-6})$ | *Score* |
|---|---|---|---|---|---|
| 0.5 | 42.18 | 27.73 | 48.09 | **7.02** | 45.14 |
| 0.4 | 42.88 | 28.17 | 49.34 | 7.52 | 46.11 |
| 0.3 | 43.58 | 28.71 | 50.92 | 8.19 | 47.25 |
| 0.2 | 44.38 | 29.36 | 52.59 | 9.06 | 48.49 |
| 0.1 | 45.31 | 30.20 | 54.40 | 11.11 | 49.86 |
| 0.05 | 45.96 | 30.69 | 56.35 | 13.32 | 51.16 |
| 0.01 | **46.10** | **31.06** | 62.28 | 25.46 | 54.19 |
| 0.005 | 45.13 | 30.96 | 64.69 | 34.76 | **54.91** |
| 0.001 | 40.41 | 30.33 | **68.82** | 79.93 | 54.61 |

**Table 4.** Effect verification of the threshold th2 in the AS strategy.

| th1 | th2 | IoU | nIoU | $P_d$ | $F_a (\times 10^{-6})$ | *Score* |
|---|---|---|---|---|---|---|
| | - | 46.10 | 31.06 | 62.28 | **25.46** | 54.19 |
| | 0.50 | **46.11** | **31.09** | 62.74 | 25.47 | 54.43 |
| | 0.45 | **46.11** | **31.09** | 63.11 | 25.62 | 54.61 |
| | 0.40 | 46.09 | 31.08 | 64.07 | 25.71 | 55.08 |
| | 0.35 | 46.01 | **31.09** | 65.24 | 26.13 | 55.63 |
| 0.01 | 0.30 | 45.89 | **31.09** | 67.00 | 26.79 | 56.45 |
| | 0.25 | 45.63 | 31.05 | 68.69 | 27.81 | 57.16 |
| | 0.20 | 45.22 | 31.00 | 70.80 | 29.54 | 58.01 |
| | 0.15 | 44.45 | 30.76 | 72.17 | 32.73 | **58.31** |
| | 0.10 | 42.68 | 30.15 | **73.45** | 40.31 | 58.07 |
| | 0.05 | 36.43 | 28.28 | 71.86 | 72.65 | 54.15 |

$P_d$ and *Score* are improved by 3.92 (from 42.18 to 46.10), 14.19 (from 48.09 to 62.28) and 9.05 (from 45.14 to 54.19), achieving the relatively optimal segmentation accuracy. Considering that the threshold th2 in the AS strategy mainly focuses on improving the detection rate, we choose the most refined segmentation scheme, that is, th1 equals 0.01, for subsequent research. Notably, our use differs from that of the original AS strategy. The main reason why th1 in this experiment can be very low (0.01) is that before adjusting the threshold th1, we use the CRF strategy in the network. The addition of the CRF strategy can refine the detected strong targets while eliminating weak targets and suppressing the generation of noise. Only when the CRF or other effective refinement post-processing strategies are introduced, the threshold th1 can be set to a very small value.

To further verify the effect of threshold th2 in the adjustable sensitivity strategy, we adjust threshold th2 when th1 is equal to 0.01. The experimental results are shown in Table 4. Notably unlike th2 in the original AS strategy, the way to introduce weak targets in threshold th2 here is obtained by processing the probability map without the CRF. The CRF suppresses the generation of weak targets, and the purpose of the threshold th2 is to introduce weak targets under a certain confidence level. From Table 4, the addition of threshold th2 will greatly improves the performance of the network. Specifically, when th2 equals 0.15, the network achieves the best comprehensive performance. Compared with the result without adding threshold th2, the detection rate is greatly improved with a slight decrease in segmentation accuracy. Its *Score* increased by 4.12 (from 54.19 to 58.31). At the same time, combining the results in Table 1 and Table 2, compared with the traditional single threshold of 0.5, the use of the AS strategy can improve the IoU, $P_d$ and *Score* by 2.27 (from 42.18 to 44.45), 24.08 (from 48.09 to 72.17), and 13.17 (from 45.14 to 58.31), respectively.

Table 5. Effects of different cropping sizes of test images

| Crop size | IoU | nIoU | $P_d$ | $F_a$ (×10$^{-6}$) | *Score* |
|---|---|---|---|---|---|
| No crop | 42.38 | 31.18 | 71.94 | 36.85 | 57.16 |
| 1024 | 42.61 | **31.26** | 71.91 | 35.38 | 57.26 |
| 512 | 43.50 | 30.98 | 71.94 | 34.90 | 57.72 |
| 256 | **44.45** | 30.76 | **72.17** | **32.73** | **58.31** |

Table 6. Performance verification of the TTA and CRF strategies.

| Settings | | | Evaluation metrics | | | | |
|---|---|---|---|---|---|---|---|
| Variants | CRF | TTA | IoU | nIoU | $P_d$ | $F_a$ (×10$^{-6}$) | *Score* |
| Our-w/o CRF | ✘ | ✔ | 0.23 | 3.71 | 37.07 | 57212.078 | - |
| Our-w/o TTA | ✔ | ✘ | 42.89 | **30.78** | 72.10 | 45.99 | 57.50 |
| Our-w/o TTA+CRF | ✘ | ✘ | 0.23 | 2.99 | 36.55 | 56761.32 | - |
| Ours(TTA+CRF) | ✔ | ✔ | **44.45** | 30.76 | **72.17** | **32.73** | **58.31** |

Table 7. Performance verification of th1 and the CRF

| th1 | CRF | IoU | nIoU | $P_d$ | $F_a$ (×10$^{-6}$) | *Score* |
|---|---|---|---|---|---|---|
| 0.5 | ✔ | 40.50 | 27.64 | 70.10 | **14.92** | 55.30 |
| 0.5 | ✘ | 42.72 | 29.45 | 70.27 | 20.84 | 56.50 |
| 0.4 | ✔ | 41.16 | 28.03 | 70.25 | 15.41 | 55.71 |
| 0.4 | ✘ | 42.10 | 29.69 | 70.85 | 39.89 | 56.48 |
| 0.01 | ✔ | **44.45** | **30.76** | **72.17** | 32.73 | **58.31** |

### 4.6   Exploration of window size in the sliding window cropping strategy

We further explore the input window size of the test images. From Table 5, when the test image is cropped to 256, both the segmentation accuracy and the detection rate achieve the best performance. This is because the network uses random cropping to 256 inputs during training, making it easier for the network to recognize test images cropped to 256. In addition, when the test images are cropped to 256 for input, the test time with the crop patch batch size set to 32 is only 0.86 times that when it is set to 1.

### 4.7   Effect verification of the TTA and CRF strategies

To verify the effect of post-processing by TTA and CRF strategies, we conduct ablation experiments on the final scheme. The thresholds th1 and th2 of the AS strategy are set to 0.01 and 0.15, respectively. From Table 6, whether removing the CRF alone or removing the TTA+CRF combination strategy, the network performance is greatly reduced. The reason is that when the threshold th1 is set to a very small value, such as 0.01, if there is no CRF to constrain the probability map, too many non-target areas will be released, causing the network to not work properly. Therefore, when th1 is set to a small value, it must be used together with the CRF strategy to ensure that the network achieves excellent performance. To further verify our conjecture, we conduct experiments with the threshold th1 increased to 0.4 and 0.5, while keeping the other conditions unchanged. The experimental results are shown in Table 7. From Table 7, when the threshold th1 is increased to 0.4 and 0.5, adding the CRF strategy worsens the network performance. However, for single-point supervision in the current framework, the combination of the small th1 and CRF strategy performs better than the case of the large th1.

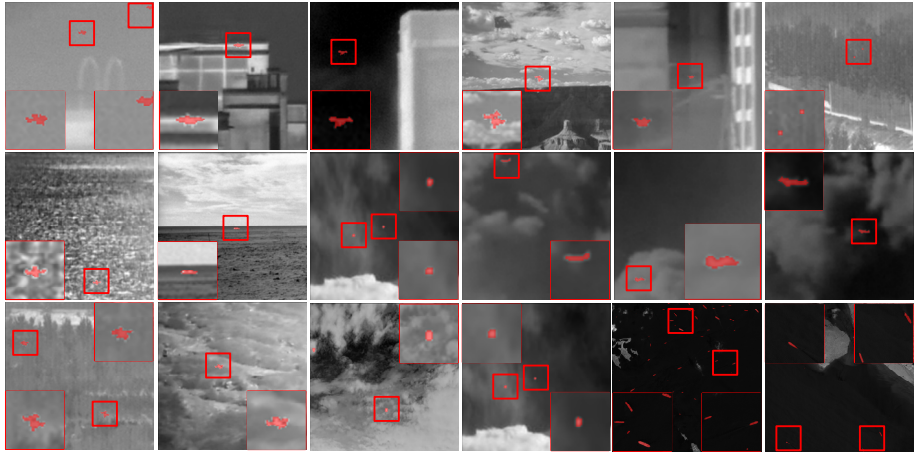

**Fig. 5.** Partial test results of the final scheme on the WP-IRSTD dataset.

To more intuitively demonstrate the detection effect of the proposed scheme, Fig. 5 shows some detection results. The refined infrared small target detection scheme with single-point supervision proposed by us achieves an excellent detection effect. For simple samples, accurate segmentation effects can be achieved, and for difficult samples, small targets can be detected to a large extent.

## 5   Conclusion

This manuscript proposes an innovative refined infrared small target detection scheme with single-point supervision, which achieves excellent detection performance. Specifically, we introduce the LESPS framework and explore the detection performance of various excellent infrared small target detection networks. GGL-Net is selected as the backbone network. At the same time, we conduct an in-depth exploration of the loss function and the test image input strategy. On the one hand, we construct and find that multi-stage loss constraints are helpful for refined detection. On the other hand, we find that reasonably sliding and cropping the test samples

14results in better performance for actual multi-size sample inputs. In addition, we propose a complete post-processing strategy. To improve the segmentation accuracy, we used a combination of TTA and CRF for post-processing during testing. On the other hand, to improve the detection rate, we introduce the AS strategy. By fully considering the various advantages of multiple detection results, some areas with low confidence are reasonably added to the fine segmentation image in the form of centroid points to significantly improve the detection rate. Extensive experimental results show that the proposed scheme achieves excellent performance.